%% file: main.tex
\documentclass{article}
\usepackage{amssymb}
\usepackage{amsmath}
\usepackage{bm}
\usepackage[preprint]{corl_2025} 

\usepackage{tabularx}
\usepackage{booktabs}
\usepackage{multirow}
\usepackage{rotating}
\usepackage{xspace}

\title{TeleOpBench: A Simulator-Centric Benchmark for Dual-Arm Dexterous Teleoperation}
\newcommand{\nickname}{TeleOpBench\xspace}

\author{
Hangyu Li$^{1,4,^{*}}$ \quad Qin Zhao $^{1,2,^{*}}$ \quad Haoran Xu$^{1,2}$ \quad Xinyu Jiang$^{7,1}$ \quad Qingwei Ben$^{1,3}$ \and
\textbf{Feiyu Jia$^{1}$} \quad \textbf{Haoyu Zhao$^{1}$} \quad \textbf{Liang Xu$^{1}$}  \quad \textbf{Jia Zeng$^{1}$}  \quad \textbf{Hanqing Wang$^{1}$}  \quad \textbf{Bo Dai$^{5,6}$} \and
\textbf{Junting Dong$^{1\dag{}}$}  \quad \textbf{Jiangmiao Pang$^{1}$} \\
$^1$Shanghai AI Laboratory \quad $^2$Zhejiang University \quad $^3$The Chinese University of Hong Kong \and $^4$ The Hong Kong University of Science and Technology (Guangzhou) \and $^5$The University of Hong Kong \quad $^6$Feeling AI \quad $^7$Tsinghua Shenzhen International Graduate School\\
$^*$Equal Contribution \quad $^{\dag{}}$Corresponding Author\\
}
    
\begin{document}
\maketitle

\vspace{-2em}
\input{sections/abstract}


\input{sections/intro}

\input{sections/related}

\input{sections/method}

\input{sections/experiments}

\input{sections/conclusion}

\clearpage
\section{Limitations}
\label{sec:limitations}
TeleOpBench presently targets upper-body teleoperation in predominantly tabletop settings. A natural next step is to build a loco-manipulation benchmark that couples dexterous arm–hand control with lower-body locomotion, thereby testing whole-body teleoperation pipelines. A second promising direction is to incorporate haptic-feedback interfaces. All modalities evaluated in this study lack tactile feedback; adding haptic would enable the assessment of finer force-controlled tasks and further broaden the benchmark’s applicability.




\bibliography{main}  

\end{document}

%% file: sections/abstract.tex
\begin{abstract}
    Teleoperation is a cornerstone of embodied‐robot learning, and bimanual dexterous teleoperation in particular provides rich demonstrations that are difficult to obtain with fully autonomous systems.  While recent studies have proposed diverse hardware pipelines—ranging from inertial motion‑capture gloves to exoskeletons and vision‑based interfaces—there is still no unified benchmark that enables fair, reproducible comparison of these systems. In this paper, we introduce \nickname{}, a simulator‑centric benchmark tailored to bimanual dexterous teleoperation.  \nickname{} contains 30 high‑fidelity task environments that span pick‑and‑place, tool use, and collaborative manipulation, covering a broad spectrum of kinematic and force‑interaction difficulty.  Within this benchmark we implement four representative teleoperation modalities—(i) MoCap, (ii) VR device, (iii) arm–hand exoskeletons, and (iv) monocular vision tracking—and evaluate them with a common protocol and metric suite. To validate that performance in simulation is predictive of real‑world behavior, we conduct mirrored experiments on a physical dual‑arm platform equipped with two 6‑DoF dexterous hands.  Across 10 held‑out tasks we observe a strong correlation between simulator and hardware performance, confirming the external validity of \nickname{}. \nickname{} establishes a common yardstick for teleoperation research and provides an extensible platform for future algorithmic and hardware innovation. The project page is \href{https://gorgeous2002.github.io/TeleOpBench/}{https://gorgeous2002.github.io/TeleOpBench/}.
\end{abstract}

%% file: sections/intro.tex
\section{Introduction}
\label{sec:intro}
\begin{figure}[htbp]
  \centering
  \includegraphics[width=0.7\textwidth]{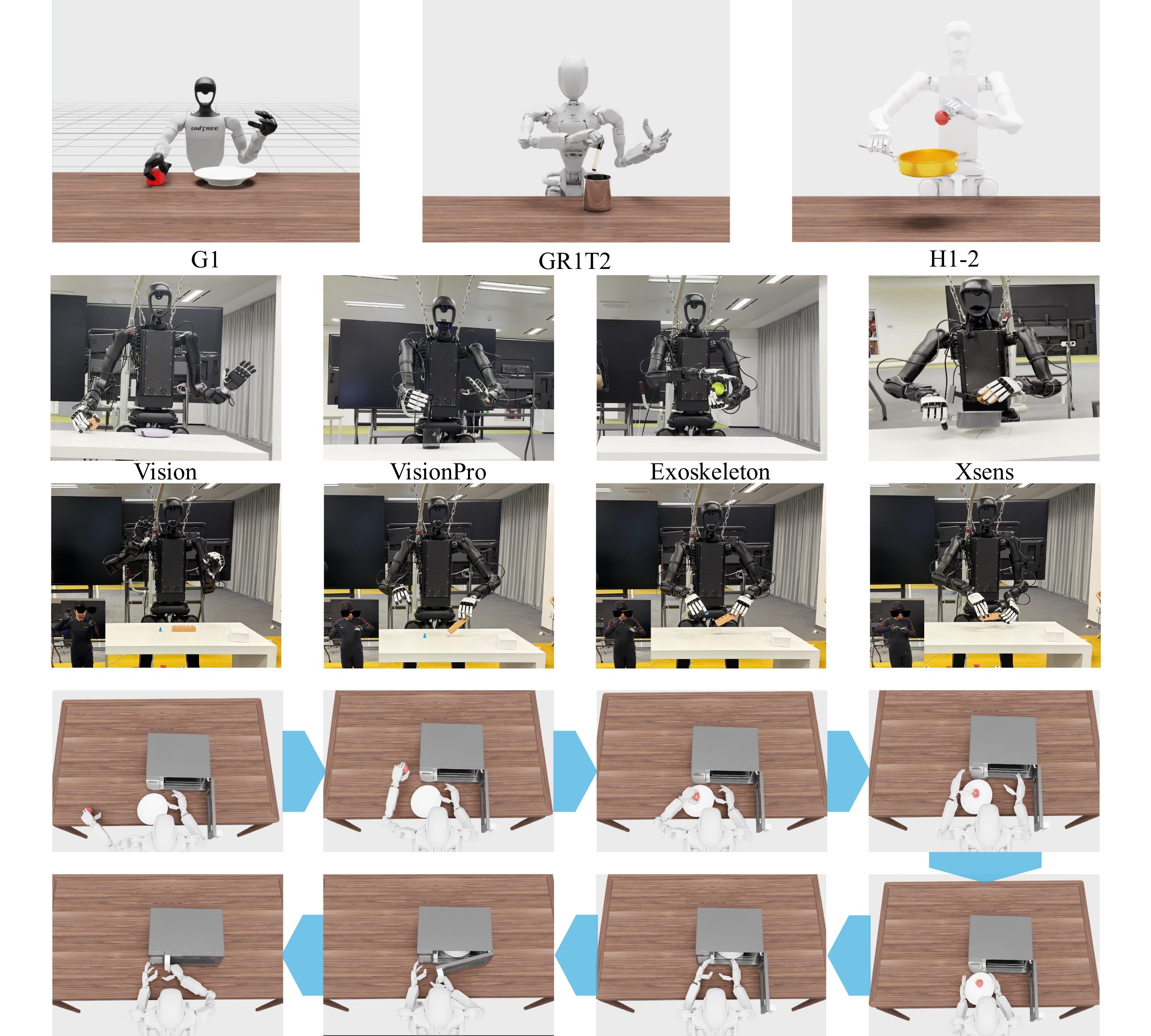}
  \caption{We present \nickname{}, a simulation-based benchmark for bimanual dexterous teleoperation, and evaluate four representative teleoperation modalities across multiple robot platforms (row 1). Real-robot experiments (row 2) demonstrate four teleoperation capabilities. Our teleoperation pipelines support fine-precision manipulation in the real world—for example, the left hand grasps a block while the right hand simultaneously inserts a smaller block (row 3)—and can execute long-horizon sequences, such as retrieving a tomato-laden plate from a microwave with the right hand and transferring the tomatoes to a table with the left (rows 4 and 5).}
  \label{fig:teasor}
  \vspace{-2em}
\end{figure}

Recent breakthroughs in robot learning \citep{chi2023diffusion,brohan2023rt,zhao2023learning} have been fueled by ever-growing repositories of human-demonstration data \citep{wang2023mimicplay, xiong2024adaptive,bahl2022human,shaw_videodex,zhu2024vision,mendonca23swim,wang2024cyberdemo}, which supply rich priors and reduce the sample complexity of learning in the real world.
Teleoperation stands out as a pivotal data-acquisition paradigm, yielding precise yet natural manipulation trajectories that are indispensable for training high-fidelity control policies.
Unlike single-arm grippers, humanoid dual-arm dexterous manipulators unlock the execution of intricate, fine-grained tasks, but their heightened kinematic complexity and need for tightly coordinated bimanual motion render teleoperation more challenging.

Recent advances in dual-arm dexterous teleoperation \citep{{lin2024learning,iyer2024open}} have showcased remarkable progress, leveraging a diverse suite of operator interfaces—from inertial and optical motion-capture setups \citep{dafarra2022icub3,cisneros2022team} to VR controllers \citep{chagas2021humanoid,holodex}, upper-body exoskeletons \citep{ramos2018humanoid,ishiguro2020bilateral,ben2025homie}, and purely vision-based trackers \citep{handa2020dexpilot,qin2023anyteleop}.
Yet, despite these impressive demonstrations, the community still lacks a standardized benchmark that would enable rigorous, fair, and comprehensive comparisons across competing approaches.
Because each system is tightly coupled to its own mix of teleoperation hardware, robot platform, and task environment, cross-method evaluation remains muddled, obscuring objective assessments of performance and ultimately hampering progress in dexterous teleoperation research.

To solve this, we introduce \nickname{}, a novel simulator-based benchmark expressly designed for fair evaluation of dual-arm teleoperation systems.
Because the simulator fixes both the robot morphology and the task environment, it eliminates the hardware and scene variability that plagues real-robot comparisons, making it uniquely suited for systematic assessment.
\nickname{} couples a broad spectrum of task environments with multiple teleoperation interfaces within a single, coherent framework.
Concretely, we provide 30 progressively harder tasks (ranging from simple cube pick-and-place to long-horizon routines such as lifting a pot lid and transferring fruit from the pot to an external dish) and the task suite can be easily extended or customized by users.
In addition, under a unified, modular interface, we implement four representative dual-arm teleoperation modalities: inertial motion capture, VR controllers, upper-body exoskeletons, and vision-only.
Researchers can effortlessly plug new teleoperation pipelines into this framework and benchmark them under exactly the same conditions, enabling truly fair comparisons.

Leveraging \nickname{}, we conduct a comprehensive evaluation of the four teleoperation modalities, reporting task-wise success rates and completion times across diverse tasks.
We further replicate nearly identical scenarios on a physical dual-arm platform and gather real-world performance metrics for each teleoperation system.
The strong correlation between simulator and hardware results confirms that \nickname{} faithfully predicts real-robot outcomes, underscoring its value as a rigorous benchmarking tool.
All code and task assets will be released open-source to foster transparent, reproducible research and to accelerate progress in dexterous teleoperation.

In summary, this paper makes the following contributions:
\begin{enumerate}
    \item We introduce a dedicated benchmark, \nickname{}, for dual-arm dexterous teleoperation, enabling rigorous, fair, and comprehensive comparisons across competing systems.
    \item We implement four representative teleoperation pipelines—motion-capture, VR controllers, upper-body exoskeletons, and vision-only within a single modular framework.
    \item Extensive experiments on both \nickname{} and a real dual-arm platform reveal a strong correlation between simulated and physical performance, substantiating the benchmark’s fidelity and practical value.
\end{enumerate}

%% file: sections/related.tex
\section{Related Work}

\noindent \textbf{Bimanual dexterous teleoperation.}
Teleoperation emerges as a crucial paradigm for acquiring robot operation data. Current teleoperation systems have evolved from grippers~\citep{{fang2023low,seo2023deep,wu2023gello,fu2024mobile}}, single-arm setups~\citep{arunachalam2023holo,qin2022one,wang2024cyberdemo} to bimanual dexterous hands~\citep{{lin2024learning,iyer2024open}}. Compared with gripper-based or single-arm set-ups, bimanual, multi-finger platforms unlock far more intricate manipulation skills, yet they also amplify the difficulty of teleoperation. To meet these new demands, researchers have recently explored a spectrum of input modalities—notably exoskeletons, motion capture (MoCap), virtual-reality (VR) devices, and purely vision-based interfaces. Exoskeleton-driven teleoperation~\citep{ramos2018humanoid,ishiguro2020bilateral,schwarz2021nimbro,fang2023low,wu2023gello,fang2024airexo,fu2024mobile,ben2025homie} removes the need for a kinematically identical master robot; joint-level matching or inverse-kinematics (IK) mapping allows operators to drive the robot with high precision and low latency. When paired with motion-sensing gloves, these systems can render truly dexterous manipulation.
VR-based approaches~\citep{lipton2017baxter,rosen2020mixed,ponomareva2021grasplook,chagas2021humanoid,holodex} employ handheld controllers or egocentric cameras to recover wrist pose and finger keypoints, which are then transformed via IK into robot joint targets, offering a cost-effective yet immersive control loop.
MoCap systems~\citep{zhao2012combining,dragan2013legibility,liu2017glove,darvish2019whole,liu2019high,liu2021semi,dafarra2022icub3,cisneros2022team,mosbach:humanoids2022}—whether inertial or optical—track full arm-hand kinematics at high frequency, achieving both high accuracy and bandwidth, but at the expense of specialized hardware and calibration effort.
Vision-only methods~\citep{handa2020dexpilot,qin2023anyteleop} estimate wrist and finger pose directly from monocular camera; although they currently lag behind MoCap in precision and update rate, they dramatically reduce deployment cost and complexity, making teleoperation more accessible.
Together, these modalities chart a rich design space for bimanual teleoperation, each balancing fidelity, latency, and affordability in different ways—trade-offs that our benchmark seeks to evaluate systematically.

\noindent \textbf{Robotics benchmark.}
A well-designed benchmark provides a standardized, reproducible and equitable environment for assessing different approaches, which substantially promotes the development of the field~\citep{deng2009imagenet,lin2014microsoft,chang2015shapenet,carreira2017quo,vuong2023open}. 
For robotics-related tasks, real-world experiments introduce significant uncertainty from hardware setups, lighting conditions and evaluation task configurations. Thus, many studies have developed simulation benchmarks as alternatives~\citep{plappert2018robotics,tassa2018deepmind,yu2019meta,zhu2020robosuite,caggiano2022myosuite,james2020rlbench,dasari2022rb2,caggiano2022myosuite,alhafez2023locomujoco,heo2023furniturebench,sferrazza2024humanoidbench}.
In particular, a growing number of simulation-based evaluation platforms have emerged for robotic reinforcement learning~\citep{openaigym2016,tunyasuvunakool2020,tassa2020dmcontrol,openaigym,yu2019meta,robosuite2020,james2020rlbench,lee2021ikea,caggiano22a} and imitation learning~\citep{memmesheimer2019simitate,al2023locomujoco,abhishek2019,rajeswaran2018}.
Comprehensive evaluation of teleoperation systems aims to quantify the performance, reliability and usability of the human operators controlling robotic platforms through various interfaces. Existing research on evaluating teleoperation systems has explored a diverse range of robotic platforms, input interfaces, task environments and realities (\textit{i.e.}, real or simulation), which makes the cross-method fair comparison and reproducibility infeasible.
Inspired by previous robotic benchmarks, we propose a simulation-centric evaluation platform named \nickname for bimanual dexterous teleoperation benchmarking, which supports various input interfaces, robot entities, and a wide range of customizable tasks.

%% file: sections/method.tex
\section{\nickname{}}

\begin{center}
\begin{figure}[t!]
  \centering
  \includegraphics[width=1\textwidth]{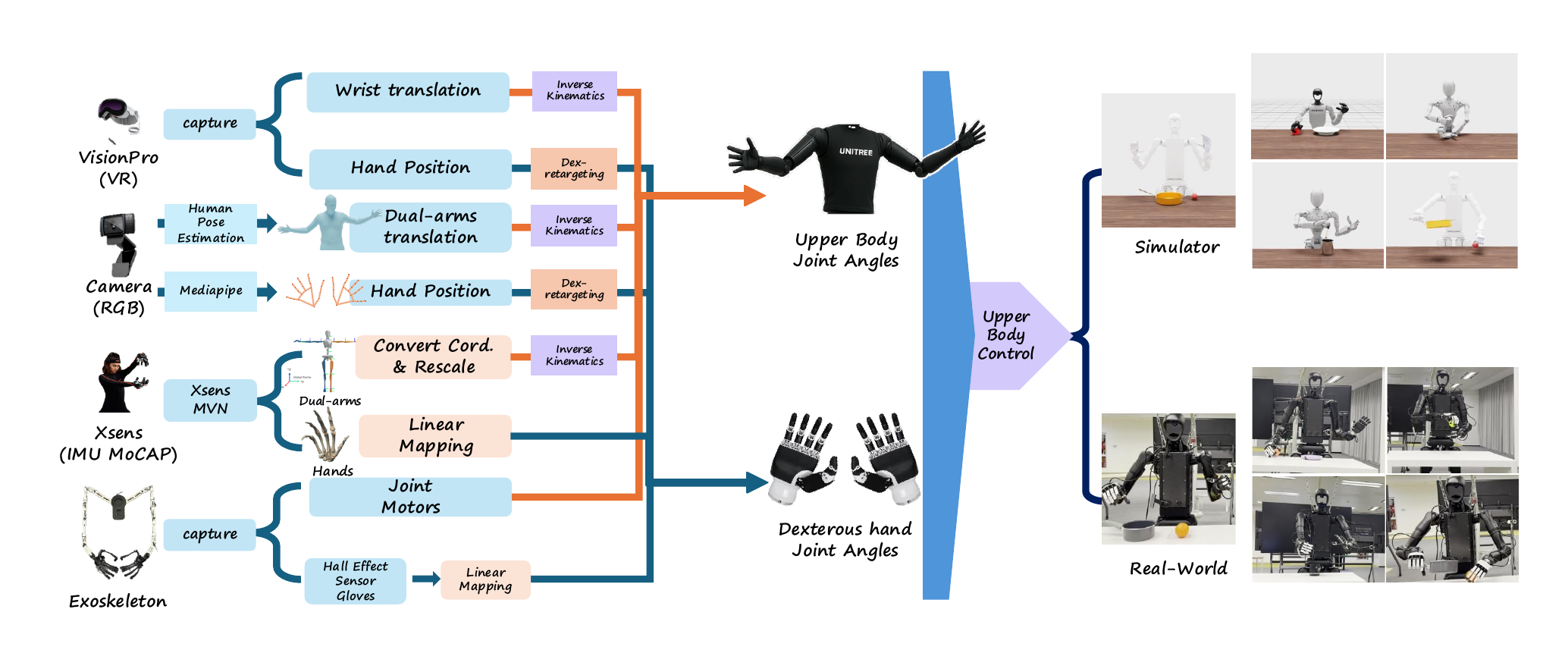}
  \vspace{-1.5em}

  \caption{The overview of the proposed \nickname{}, where we unify four operator interfaces in both simulation and real-world realities for dual-arm dexterous teleoperation.}
  \label{fig:pipeline}
\vspace{-1.5em}
\end{figure}
\end{center}

We introduce \nickname{}, a simulator-based benchmark purpose-built for impartial evaluation of dual-arm teleoperation systems.
Figure \ref{fig:pipeline} shows an overview of \nickname{}.
Leveraging the simulator’s controllability, \nickname{} provides 30 task environments spanning a wide difficulty spectrum—from elementary cube pick-and-place to long-horizon routines such as lifting a pot lid and transferring fruit from the pot to an external dish (Section \ref{sec:sim}).
Furthermore, four representative teleoperation pipelines—inertial motion capture, VR controllers, upper-body exoskeletons, and vision-only hand tracking—are implemented under a unified, modular interface and instantiated on three different dual-arm robots (Section \ref{sec:teleop}).

\vspace{-1.5em}
\subsection{Task environments} \label{sec:sim}  

We employ NVIDIA Isaac Sim as our simulation platform because its high-performance PhysX engine and photorealistic renderer enable the construction of environments that closely approximate real-world conditions. Each scene features a humanoid robot fitted with bimanual dexterous hands and the task-relevant objects; operators are instructed to execute the required manipulations exactly as specified. For every trial, we record both task success and completion time, which together constitute our primary performance metrics.

\textbf{Humanoid robots.} For a comprehensive hardware evaluation, we employ three commercially available humanoid platforms—Unitree H1-2, Fourier GR1-T2, and Unitree G1. Unitree H1-2 is a full-size humanoid with 7 DoF in each arm. The base model is equipped with Inspire Dexterous Hands with five articulated fingers and 6 DoF for fine manipulation. Fourier GR1-T2 matches2 in overall stature and arm kinematics (7 DoF)DoF),t ships with Fourier’s native dexterous hands, which provide five fingers and 6 DoF. Unitree G1 adopts a far more compact form factor and is equipped with lightweight three-finger hands offering 4 DoF. The trio spans a meaningful range of scale and hand design, enabling a systematic assessment of teleoperation across diverse humanoid robots.

\textbf{Task setting.} \nickname{} offers a comprehensive, multi-tiered suite of 30 bimanual dexterous-manipulation tasks. The tasks are hierarchically organized by complexity—e.g., whether they require coordinated two-hand interaction or long-horizon sequencing—so that both coarse- and fine-grained teleoperation modalities can be evaluated in an appropriately graduated setting. This progression is essential: if all tasks were uniformly difficult, lower-precision interfaces such as monocular-vision tracking would fail wholesale; if all tasks were trivial, higher-precision methods such as inertial MoCap would be indistinguishable from less capable alternatives.

To facilitate customization and encourage community contributions, we provide a modular, extensible asset library with flexible APIs that let researchers instantiate new scenarios or tune task difficulty with minimal effort. Every task includes explicit success criteria, ensuring that results are assessed under consistent, objective standards. In simulation, we replicate real-world physics as closely as possible to match object masses, friction coefficients, and other dynamics so that each scenario remains physically and pragmatically faithful. A complete list of tasks and their configurations is shown in Table \ref{tb:task}.

\begin{table}[t]
\centering
\setlength{\tabcolsep}{4pt}
\renewcommand{\arraystretch}{1.2}
\normalsize
\resizebox{\columnwidth}{!}{%
\begin{tabular}{@{}>{\arraybackslash}p{3.2cm} 
                >{\arraybackslash}p{6.5cm} 
                >{\arraybackslash}p{3.2cm} 
                >{\arraybackslash}p{7.5cm}@{}}
\toprule
\textbf{Task Name} & \textbf{Completion Criteria} & \textbf{Task Name} & \textbf{Completion Criteria} \\
\midrule
push\_cube & push the cube into the left blue area & pot\_bimanual & lift the pot with both hands \\
pick\_cube & pick up and place the cube down & pot\_tomato & lift the pot and add tomato into it \\
pick\_place\_cube & pick up the cube and place it on the plate & pot\_tray & place the pot onto the tray with both hands \\
uprear\_cup & pick up the cup and place it upright & stack\_boxes & lift the box and stack it onto another box \\
ball\_trashcan & pick up the ball and place it into the trashcan & pan\_hearth & open the lid and place the pan on the hearth \\
rotate\_faucet & rotate the faucet 90 degrees & tidyup\_table & place objects into the basket in order \\
rotate\_hearth & press and turn the hearth knob & pour\_water & pour water from the kettle into the cup \\
open\_microwave & pull open the microwave door & pot\_tomato\_out & open the pot and remove the tomato from it \\
close\_microwave & close the microwave door & plate\_oven & open the oven, place the plate, close it \\
open\_drawer & pull open the drawer & pot\_tomato\_plate & open the lid and place the tomato on the plate \\
close\_drawer & close the drawer & pen\_brushpot & pick up the pen and place it into the container \\
lift\_mug & lift the lid of the mug & drawer\_book & open the drawer and place the books inside \\
open\_laptop & open the laptop & bread\_toaster & place the bread into the toaster and press the button \\
ball\_mug & pick up the ball and place it into the mug & stack\_toyblocks & assemble the toy blocks in sequence \\
ball\_bimanual & pass the ball from one hand to another & twist\_bottle\_cap & pick up the bottle and twist the cap on \\
\bottomrule
\end{tabular}
}
\vspace{0.5em}
\caption{The list of task names and completion criteria of \nickname.}
\vspace{-1.5em}
\label{tb:task}
\end{table}

\textbf{Observation.}
Our simulation framework records a rich set of observations for every teleoperation episode, enabling downstream imitation-learning studies and further amplifying the value of \nickname{}. The logged data include
(a) robot-state vectors—joint positions, angles, velocities, and related kinematics;
(b) camera streams—RGB images from a head-mounted, first-person camera and a fixed third-person camera facing the workbench; and
(c) task-level environment metadata—precise object positions and orientations.


\subsection{Modular Teleoperation interface} \label{sec:teleop}

We implement four representative teleoperation pipelines—monocular vision, MoCap, VR, and exoskeleton— under a unified, modular interface.

\subsubsection{Vision-based}
Unlike prior vision-based teleoperation methods, we decouple arm-and-wrist pose estimation from hand-keypoint estimation, resulting in a more robust and higher-precision visual interface. Concretely, our vision pipeline consists of three core modules: human body parameter estimation and scale, upper-body limb motion control, and hand control.

\textbf{Human-body parameter estimation and scaling.}
To reduce the sensitivity of teleoperation accuracy to inter-subject anthropometric differences, we follow the philosophy of PHC \citep{luo2023phc} and build a keypoint-constrained parameter–scaling model. In contrast to PHC, we solve the scaling parameters only once under a neutral T-pose, which is sufficient for subsequent sessions of the same operator. We select four anatomically consistent landmarks—pelvis, shoulders, wrists, and head—on both the human operator and the robot. The body parameter is obtained by minimizing the keypoint alignment error, formulated as:
\begin{equation}
\mathbf{\beta^*}= \arg\min_{\beta \in \mathbb{R}^{10}} \sum_{l} \| f_{\text{SMPL}}^l(\beta,\theta_0) - R^l \|_2^2, 
\end{equation}
where $f_{\text{SMPL}}^l(\beta,\theta)$ denotes the 3D position of the $l$-th human landmark generated by the parametric human model SMPL \citep{loper2023smpl} from shape parameter $\beta$ and pose parameter $\theta$. The vector $\theta_0$ represents the T-pose and $R^l$ denotes the 3D position of the $l$-th landmark on the robot.
We employ gradient-based optimization to obtain the optimal body parameter $\beta^*$. Under $\beta^*$, we calculate the scaling factors between each human joint link and its corresponding robotic counterpart, thereby deriving a set of optimal scale parameters $s^*$.

\textbf{Upper-body limb motion control.} We utilize SMPLer-X~\citep{cai2023smplerx} to capture the teleoperator's SMPL pose parameter, then compute joint positions under the optimized body parameter $\beta^*$. These positions are scaled by the derived factors $s^*$ to obtain robotic translations, followed by PINK~\citep{pink} based inverse kinematics solving to derive all DoF except finger joints.

\textbf{Hand control.} We provide two control schemes.
Scheme 1: Directly use the finger rotations from SMPLer-X to calculate the corresponding Euler angles, then set the robotic hand's DoF values based on these angles.
Scheme 2: Utilize MediaPipe to capture finger keypoint positions, then apply vector optimizers via Dex-Retargeting~\citep{qin2023anyteleop}—a highly versatile and computationally efficient motion retargeting library—achieving significantly improved performance. We adopt Scheme 2 throughout all experiments.
Finally, we implement a Kalman filter to smooth the robot's DoF, significantly reducing jitter-related instability in motion execution.

\subsubsection{MoCap-based}
\textbf{Hardware.} We employ the Xsens MVN system~\citep{roetenberg2009xsens} as our motion capture solution.  For capturing limb movements, the core sensor assembly includes seventeen IMUs strategically attached to corresponding human body segments. Furthermore, Xsens Metagloves by Manus are utilized to precisely capture intricate hand motions through a sophisticated hand model, providing 20 degrees of freedom (DOFs) for each hand individually. Upon wearing the Xsens suit and completing the calibration process, data capturing the positions and orientations of the seventeen body segments, as well as detailed finger joint angles, is directly accessible through the MVN system.

\textbf{Arm control.} The raw data obtained from the Xsens system is initially defined within its proprietary global coordinate system. Thus, the first necessary step involves transforming this data into the robot's coordinate system. Within our teleoperation setup, the robot's lower body is immobilized; hence, we define the robot's coordinate reference frame at the pelvis joint, with the robot's forward-facing direction aligned with the positive X-axis and the vertical direction aligned with the Z-axis, following a right-handed coordinate system convention. To address the inherent skeletal differences between the humanoid robot and the human operator—which could lead to significant discrepancies in motion if directly used in IK—a joint-specific rescaling method is introduced. This rescaling approach calculates and updates scaling parameters in real-time upon receiving the initial raw data. Subsequently, it adjusts joint lengths individually, converting the upper limb and arm joint coordinates accurately into the robot's coordinate frame. Finally, we compute the robot’s joint poses using Closed-Loop Inverse Kinematics (CLIK), ensuring precise and robust teleoperated control.





\textbf{Hand control.} The Manus glove provides detailed axis-angle data representing 20 degrees of freedom per hand. This data encompasses the flexion and abduction/adduction movements of the metacarpophalangeal (MCP) joints (connecting each finger to the palm), as well as the flexion of the proximal interphalangeal (PIP) joints (the intermediate joints of each finger) and the distal interphalangeal (DIP) joints (nearest to each fingertip). The captured finger motion data from the Manus gloves is subsequently mapped to the joint angle constraints defined by the robotic dexterous hand (dexhand), ensuring accurate and precise finger control.
\subsubsection{VR-based}
The VR-based teleoperation system consists of upper-body limb motion control and hand control.

\textbf{Upper-body limb motion control.} We utilize the Apple VisionPro for hand, wrist, and head tracking. The tracking adheres to the OpenXR coordinate system. The wrist and head poses are first transformed into the robot's coordinate frame, and the wrist offset relative to the head is subsequently converted into an offset relative to the pelvis. We exclusively feed the wrist translation data to the IK algorithm based on Pink~\citep{pink2024}, which computes all degrees of freedom except finger joints.

\textbf{Hand control.} To enhance manual dexterity across different teleoperators, we measure the distal phalanx lengths of each operator's fingers and scale them proportionally to match the corresponding robotic finger segments, resulting in a scaling factor $s^* \in \mathbb{R}^{5}$.
Subsequently, following OpenTelevision \citep{cheng2024open}, we employ vector-based optimizers to generate robot-hand joint commands via the dexterous-retargeting framework of AnyTeleop \citep{qin2023anyteleop}.

\subsubsection{Exoskeleton-based}

We propose a framework for designing isomorphic exoskeleton systems tailored to diverse humanoid platforms, enabling high-precision teleoperation across simulated and physical environments. Building on principles from HOMIE~\citep{ben2025homie}, each exoskeleton is customized to replicate the kinematic structure of its target humanoid’s upper body, utilizing servo-driven joints to synchronize human operator movements with robotic counterparts in real time. Integrated motion-sensing gloves equipped with Hall-effect sensors provide 15 degrees of freedom (DoF) per-hand tracking for dexterous manipulation. By directly mapping operator kinematics to the humanoid’s joints—bypassing inverse kinematics (IK) approximations—our platform-specific exoskeletons eliminate algorithmic errors while enhancing operational bandwidth and positional accuracy. 

%% file: sections/experiments.tex
\section{Experiments}
\label{sec:result}

\begin{figure}[htbp]
  \centering
  \includegraphics[width=\textwidth]{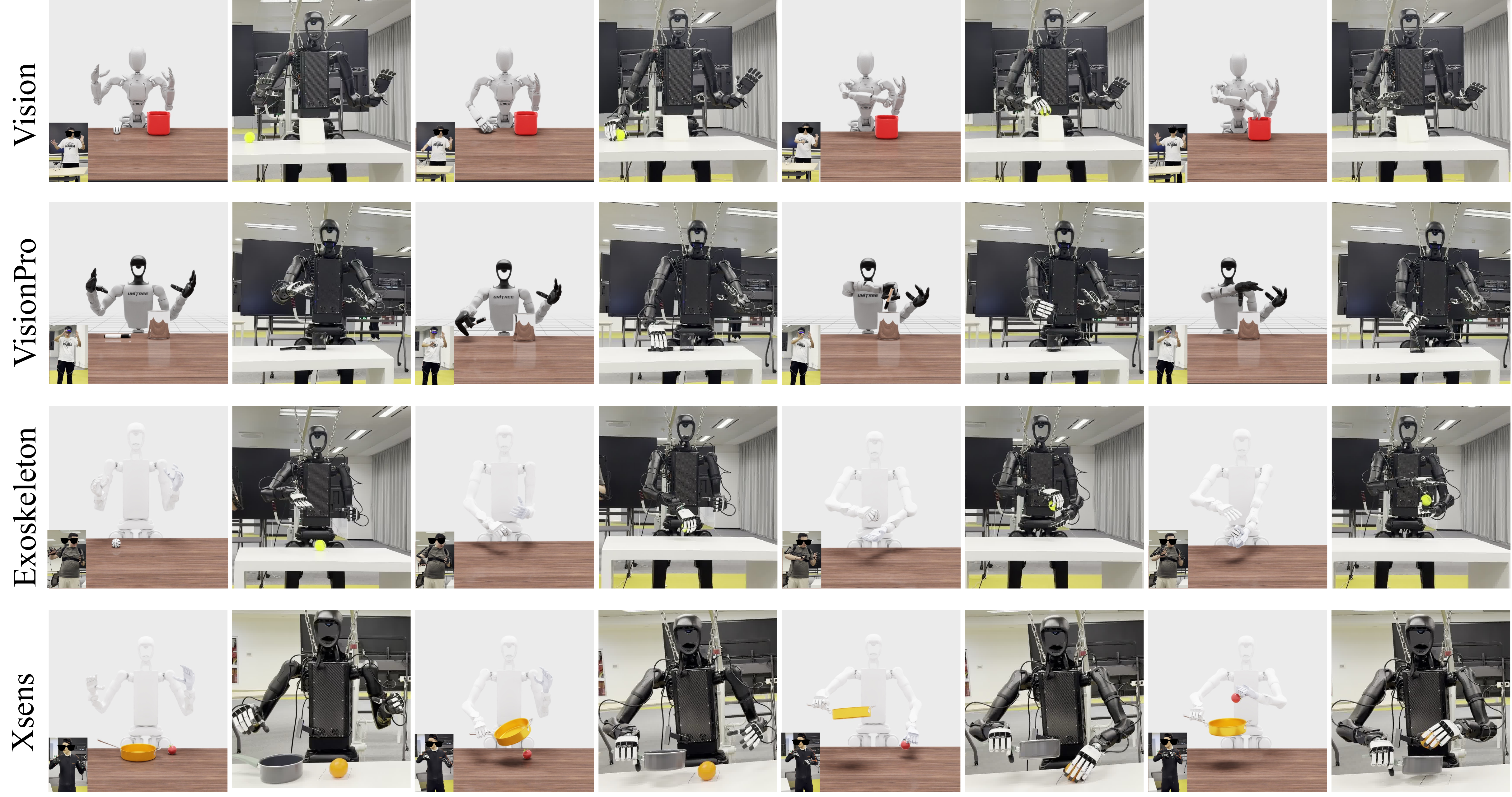}
  \caption{From top to bottom, we illustrate the four teleoperation modalities executing the following tasks: ball\_trashcan, pen\_brushpot, ball\_bimanual, and pot\_bimanual.}
  \label{fig:process}
\end{figure}



In this section, we evaluate the effectiveness of TeleOpBench. First, we showcase the performance of four teleoperation schemes across both simulated and real-world tasks. We then carry out a systematic comparison in the simulator and analogous task settings on physical hardware.


\textbf{Qualitative results.} Figure \ref{fig:process} presents qualitative results of the four teleoperation modalities across both simulated and real-world task settings. Our vision-based system achieves a high success rate in grasping tasks, e.g., accurately placing a ball into a trash bin. The VR-based system successfully places a slender pen into a pen holder. The exoskeleton-based system allows the right hand to pick up a ball and transfer it to the left hand for a stable grasp. The Xsens motion-capture system supports high-precision manipulation—e.g., picking up a slender pen into a pen holder.

\begin{table}[htbp]
\centering
\resizebox{0.7\textwidth}{!}{%
\begin{tabular}{@{}l*{8}{c}@{}}
\toprule
\multirow{2}{*}{Task} & 
\multicolumn{2}{c}{\textbf{Vision-Based}} & 
\multicolumn{2}{c}{\textbf{VR-Based}} & 
\multicolumn{2}{c}{\textbf{Exoskeleton}} & 
\multicolumn{2}{c}{\textbf{Xsens}} \\
\cmidrule(lr){2-3} \cmidrule(lr){4-5} \cmidrule(lr){6-7} \cmidrule(lr){8-9}
& Succ (\%) & Time (s) & Succ (\%) & Time (s) & Succ (\%) & Time (s) & Succ (\%) & Time (s) \\
\midrule
1  &80 &13.64 &100 &15.32 &90 &16.42 &100 &6.32 \\
2  &100 &34.66 &100 &15.54 &100 &12.69 &100 &7.34 \\
3  &80 &33.00 &100 &12.46 &90 &20.28 &100 &7.87 \\
4  &40 &56.50 &80 &21.67 &80 &16.91 &90 &11.19 \\
5  &70 &35.96 &100 &12.51 &100 &15.48 &100 &14.52 \\
6  &60 &52.75 &100 &15.13 &90 &17.86 &100 &9.97 \\
7  &0 &-- &0 &-- &80 &21.06 &100 &14.70 \\
8  &50 &36.64 &100 &9.62 &100 &7.85 &90 &16.36 \\
9  &10 &24.87 &90 &41.72 &80 &37.86 &100 &11.38 \\
10 &0 &-- &70 &57.32 &80 &23.92 &100 &12.63 \\

\bottomrule
\end{tabular}%
}
\vspace{0.5em}

\caption{Performance comparison of teleoperation systems across tasks in simulation.}  \label{tab:sim_comparison}
\vspace{-2em}
\end{table}

\textbf{Quantitative results in simulation tasks.}
We select ten representative tasks of varying difficulty from the TeleOpBench: (1) push\_cube, (2) pich\_cube, (3) pick\_place\_cube, (4) uprear\_cup, (5) ball\_trashcan, (6) ball\_mug, (7) ball\_bimanual, (8) pot\_bimanual, (9) pot\_tomato\_plate, and (10) pen\_brushpot. Full task descriptions refer to Table \ref{tb:task}. A user study involving four participants was conducted; Task-level success rates and completion times are summarized quantitatively in Table \ref{tab:sim_comparison}.













For vision-based methods, monocular camera keeps the setup simple, but low frame-rate, coarse wrist-orientation estimates, and occlusion limit it to easy tasks, yielding the largest completion times and poor success on Tasks 4 and 7. For VR, accurate wrist/hand tracking gives strong grasps, yet Task 7 fails entirely because hand-over-hand occlusion breaks pose estimation. The exoskeleton-based method, with a kinematically aligned design and direct DoF mapping to the robot, delivers smooth control and performs well in most tasks. However, due to limited capability in lateral elbow movement, it shows increased time consumption in Task 1 Push the cube. The Xsens-based method excels in both smoothness and motion precision. It completes the tasks accurately and typically with the least time cost. However, it is also the most expensive among the four teleoperation systems.

\begin{table}[htbp]
\centering
\resizebox{0.7\textwidth}{!}{%
\begin{tabular}{@{}l*{8}{c}@{}}
\toprule
\multirow{2}{*}{Task} & 
\multicolumn{2}{c}{\textbf{Vision-Based}} & 
\multicolumn{2}{c}{\textbf{VR-Based}} & 
\multicolumn{2}{c}{\textbf{Exoskeleton}} & 
\multicolumn{2}{c}{\textbf{Xsens}} \\
\cmidrule(lr){2-3} \cmidrule(lr){4-5} \cmidrule(lr){6-7} \cmidrule(lr){8-9}
& Succ (\%) & Time (s) & Succ (\%) & Time (s) & Succ (\%) & Time (s) & Succ (\%) & Time (s) \\
\midrule
1  &100 &14.41 &100 &15.29 &90 &18.47 &100 &10.44 \\
2  &70 &30.79 &100 &9.82 &100 &9.77 &100 &8.31 \\
3  &80 &14.95 &100 &10.16 &100 &8.48 &100 &6.12 \\
4  &40 &24.79 &70 &14.32 &80 &15.22 &90 &11.33 \\
5  &60 &23.11 &100 &13.57 &80 &12.43 &100 &6.91 \\
6  &20 &26.21 &90 &13.86 &60 &16.75 &100 &8.18 \\
7  &0 &-- &0 &-- &70 &24.82 &90 &12.90 \\
8  &40 &26.34 &100 &11.18 &100 &5.49 &100 &12.02 \\
9  &10 &53.34 &90 &36.32 &80 &22.43 &100 &17.97 \\
10 &0 &-- &80 &24.31 &70 &27.49 &100 &16.47 \\

\bottomrule
\end{tabular}%
}
\vspace{0.5em}
\caption{Performance comparison of teleoperation systems across tasks in real world.} \label{tab:real_comparison}
\end{table}

\vspace{-1.5em}

\begin{figure}[htbp]
  \centering
  \includegraphics[width=0.8\textwidth]{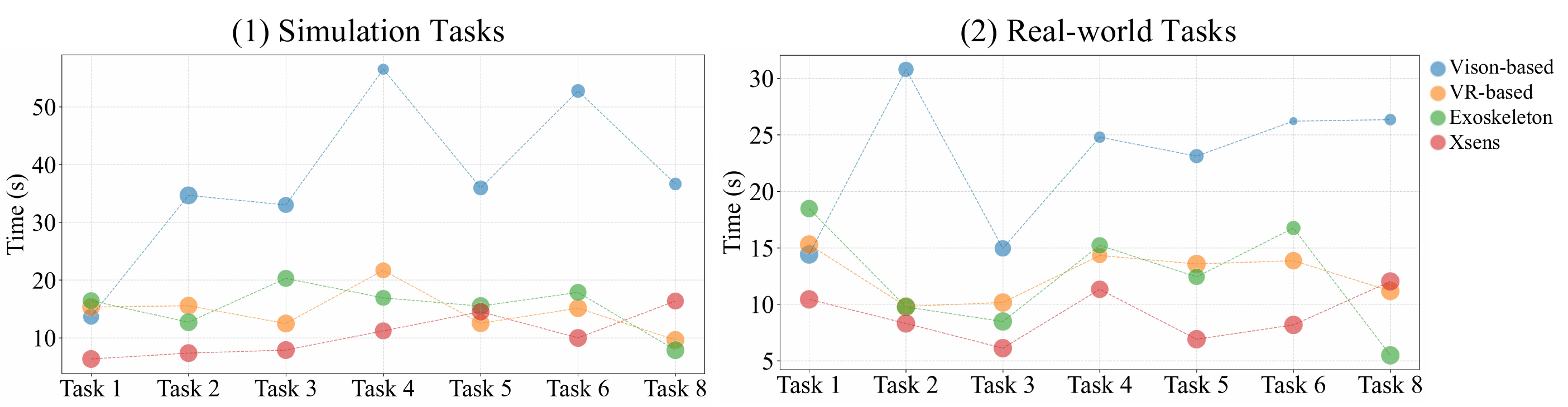}
  \caption{Completion-time and success rate curves for simulation and real world. The size of each circle reflects the corresponding task success rate.}
  \label{fig:sim2real}
\end{figure}

\textbf{Quantitative results in physical tasks.} We reproduce the task suite on physical robots and evaluate all four teleoperation pipelines with the identical metric suite; the resulting quantitative scores are summarized in Table \ref{tab:real_comparison}. Figure \ref{fig:sim2real} presents completion-time curves for simulation and real world. Note that Tasks with one teleoperation success rate below 20\% are excluded from the plotted curves to ensure the reliability of the curves. The two domains exhibit a strong positive correlation: the vision-tracking interface consistently requires the longest execution time (blue curve), the inertial-MoCap pipeline is the fastest (red curve), and the VR and exoskeleton interfaces cluster in between. This close alignment between simulated and real-world performance confirms that TeleOpBench reliably predicts practical outcomes and therefore offers substantial utility to the community.

\vspace{-1.5em}



%% file: sections/conclusion.tex
\section{Conclusion}
\label{sec:conclusion}
We present \nickname{}, a simulator-centric benchmark for bimanual dexterous teleoperation, providing a fair, reproducible platform for cross-system comparison. \nickname{} contains 30 high-fidelity task environments spanning a broad spectrum of difficulty. Within this suite, we implement four representative teleoperation modalities in a unified, modular framework. Extensive experiments in both simulation and on physical hardware reveal a strong correlation between simulated and real-world performance, validating the benchmark’s external fidelity and underscoring its practical value for future research.